\documentclass[letterpaper]{article} 
\usepackage{aaai25}  
\usepackage{times}  
\usepackage{helvet}  
\usepackage{courier}  
\usepackage[hyphens]{url}  
\usepackage{graphicx} 
\def\UrlFont{\rm}  
\usepackage{natbib}  
\usepackage{caption} 
\usepackage{algorithm}
\usepackage{algpseudocode}
\usepackage{enumitem}
\usepackage{amsmath}
\usepackage{mdframed}

\usepackage{newfloat}
\usepackage{listings}
\DeclareCaptionStyle{ruled}{labelfont=normalfont,labelsep=colon,strut=off} 
\floatstyle{ruled}
\newfloat{listing}{tb}{lst}{}
\floatname{listing}{Listing}
\title{MISCON: A Mission-Driven Conversational Consultant for Pre-Venture Entrepreneurs in Food Deserts}
\author{
    Subhasis Dasgupta\textsuperscript{\rm 1}, Hans Taparia\textsuperscript{\rm 2}, Laura Schmidt\textsuperscript{\rm 3} and Amarnath Gupta\textsuperscript{\rm 1}
}
\affiliations{
    \textsuperscript{\rm 1}San Diego Supercomputer Center, University of California, San Diego, USA\\
    \textsuperscript{\rm 2} NYU Stern School of Business, USA\\
    \textsuperscript{\rm 3}School of Medicine, University of California, San Francisco, USA\\
    sudasgupta@ucsd.edu, ht29@stern.nyu.edu, laura.schmidt@ucsf.edu, a1gupta@ucsd.edu 
}

\usepackage{bibentry}

\begin{document}

\maketitle

\begin{abstract}
This work-in-progress report describes MISCON, a conversational consultant being developed for a public mission project called NOURISH. With MISCON, aspiring small business owners in a food-insecure region and their advisors in Community-based organizations would be able to get information, recommendation and analysis regarding setting up food businesses. MISCON conversations are modeled as state machine that uses a heterogeneous knowledge graph as well as several analytical tools and services including a variety of LLMs. In this short report, we present the functional architecture and some design considerations behind MISCON.
\end{abstract}

%

\section{Introduction} \label{sec:intro}
This is a work in progress report on a \textit{conversational consultant} we are developing under a Convergence Accelerator project called NOURISH, funded by the National Science Foundation (NSF) and the US Department of Agriculture (USDA). The conversational consultant, broadly defined as an AI agent that converses with a human user within a specific domain of expertise, is a part of a knowledge-based platform being developed to interact with current and prospective food business owners with small capital and living in food deserts. 

\subsection{Public Mission}
Food deserts are areas where residents have limited access to affordable and nutritious food, particularly fresh fruits, vegetables, and whole foods. This typically means that there are no affordable grocery stores within a reasonable distance (defined by USDA as more than 1 mile in urban areas or 10 miles in rural areas). Not surprisingly, food desert regions in the US strongly overlap with low-income areas where stores sell cheaper, unhealthy, ultra-processed food, and the occurrence of obesity, diabetes, and cardiovascular disorders are significantly high.

We conducted more than 200 interviews with individuals and organizations across the Food System Network in San Diego County - covering farmers to current and aspiring business owners, community-based organizations (e.g., CDC - Community Development Corporations) to Government-funded nonprofit organizations (e.g., SBDC - Small Business Development Corporations), Federal Government organizations (e.g., SBA) to local Government organizations (e.g., City of San Diego) and identified that the NOURISH platform can needs to fill in the following unmet needs.
\begin{enumerate}
    \item Although there is a strong demand for fresh food and a desire to start small local food businesses, there is a significant \textit{\textbf{information asymmetry}} between these small business aspirants and large food chains. Today, prospective entrepreneurs do not have comprehensive and easy-to-access resources that would help navigate the complex landscape of market demand, competition, cost of business, funding opportunities, and regulatory requirements. 
    \begin{enumerate}
        \item Many of the information, such as the municipal codes of cities, is publicly available, but it is difficult to discover and compile together by ``Googling it''.
        \item Some information, for example, on legal and government funding, is written in a way that is not intelligible to our constituency of users due to their level of education and linguistic proficiency.
    \end{enumerate}
    \item Currently, their go-to places are institutions like CDCs and SBDCs which are deeply caring, well connected but understaffed, and often rely on the personal knowledge and network of individual advisors.
    \item A very large segment of the population that seeks help from the above institutions need to go through a long \textit{business ideation process} to determine the specific business and product lines that would work for them.
\end{enumerate}

\subsection{Desiderata}

MISCON, the conversation consultant we are developing is based on the following basic premises. 
\begin{itemize}
    \item It will confine itself to the domain of food-related businesses, but within this domain, it will be based on a knowledge graph that will be regularly updated as new information resources are added to the system. The knowledge graph would serve as an \textit{\textbf{integration point for heterogeneous data}} obtained from different sources of information.
    \item As part of the conversation, it will \textit{\textbf{elicit}} user-specific information that will be stored securely and used to carry forward the conversation.
    \item At any point where the system determines that the user does not have adequate information to make a decision or to produce more meaningful examples, it \textit{\textbf{proactively}} offers relevant information for the user to consume as an aid to carry out the conversation \cite{guo2024pcqpr}.
    \item Following the guidelines offered by our SBDC and CDC partners, the conversation process will aim to take the user closer to business creation. To this end, it will utilize the same set of \textit{\textbf{business milestones}} that these introductions use for their clients. 
    \item Regardless of the complexity of the content of the knowledge, the system-generated utterances in the conversation should be \textit{\textbf{user-aware}}, i.e., in accordance with the educational level of the user (obtained when the user registers with the system).
\end{itemize}

The purpose of this paper is to present the basic design and a concrete example of the conversation process. We assume that the knowledge graph, together with the milestone descriptions, is already available and can be accessed by the conversation system via internal queries presented later.

\section{Modeling Mission-Driven Conversations}

Recent surveys on conversational (dialogue) systems \cite{ni2023recent,algherairy2024review} classify them into two broad categories: open domain and task-oriented. Open domain systems are primarily data driven, where the data can come from the user's current utterance and context incrementally constructed via the conversation history, as well as from a knowledge base of factual information maintained by the system. Task-oriented systems, on the other hand, generate conversations around a predefined narrowly scoped task (e.g., travel planning) and take the user through a series of steps to accomplish each subtask (e.g., booking an airline ticket) of the primary task. For a problem like ours, neither of these models holds exactly because of the variability of the circumstances from which the user comes from. A user without a well-defined business idea would like to have more of a knowledge-guided, open-\textit{ish} domain conversation with significant drifts, while a user who knows the business type and products but is looking for a suitable location and regulations will need a more task-oriented conversation. In addition, conversation questions and responses will be based on \textit{ milestones}, including the user's state (e.g., does the user have a work permit in the US) and business mission (e.g., does the user have a registered business already). We use the term \textit{mission driven conversation} to highlight that the conversation is a complex, multistage goal (mission) often involving multiple interdependent milestones - it allows for exploration, variability, and evolving user needs. We also consider that while the conversation system uses a knowledge base (in our case, a knowledge graph), system's stored knowledge will always be incomplete for a specific user's needs and the gaps must be elicited as part of the conversation \cite{tan2023guiding}.

\subsection{An Example Conversation} \label{sec:conversation}
\begin{figure*}[t] \small
\begin{mdframed}
\noindent\textbf{Turn 1: Initial Inquiry}\\
\noindent\textit{User: "I'm interested in starting a bakery in San Ysidro, San Diego County, California. What do I need to know?"}

\noindent\texttt{System:} "Starting a bakery in San Ysidro requires understanding the local market, demographics, regulatory requirements, and financial readiness. Which aspect would you like to explore first?"

\noindent\textbf{Turn 2: Exploring the Market} \\
\noindent\textit{User: "Let’s start with the market. What should I know about San Ysidro?"}

\noindent\texttt{System:} "San Ysidro has a diverse population with lower median income levels and a spending focus on essentials and affordability. Artisan bakeries may face challenges due to limited demand for premium products. Have you considered aligning your bakery concept with local preferences?"

\noindent\textbf{Turn 3: Adjusting the Concept}\\
\noindent\textit{User: "That’s a good point. What adjustments would you recommend?"}

\noindent\texttt{System:} "In San Ysidro, you might focus on affordable baked goods, like pan dulce or other culturally relevant options. If this aligns with your skills, we can explore product localization further. Does this fit your vision?"

\noindent\textbf{Turn 4: User Provides Specific Feedback}\\
\noindent\textit{User: "Yes, I could include traditional Mexican baked goods like pan dulce. I’d also like to know how this impacts my budget."}

\noindent\texttt{System:} "Incorporating traditional baked goods may reduce your costs for premium ingredients while improving local appeal. Would you like help estimating your ingredient costs or refining your overall budget?"

\noindent\textbf{Turn 5: Dependency on Permits}\\
\noindent\textit{User: "Let’s refine the budget. I estimate around \$120,000 in startup costs, but I’m unsure about permit fees."}

\noindent\texttt{System:} "Permit fees include costs for a business license, health permits, and a seller’s permit. These can total \$1,000 to \$5,000 depending on your bakery size and location. Shall we prioritize the health permit since it has layout dependencies?"

\noindent\textbf{Turn 6: Shifting Intents}\\
\noindent\textit{User: "Sure, let’s focus on the health permit. My layout includes areas for baking, cooling, and retail space."}

\noindent\texttt{System:} "Your layout looks comprehensive. When submitting your health permit application, ensure it includes refrigeration details for cooling and customer flow in the retail area. Would you like help preparing the submission?"

\noindent\textbf{Turn 7: Dependency on Financial Readiness}\\
\noindent\textit{User: "Yes, and I’ve secured a loan for \$80,000. I’d like to understand how this affects my timeline."}

\noindent\texttt{System:} "With \$80,000 in loan funding, you might prioritize securing permits and purchasing essential equipment. Shall we map a timeline based on these milestones?"

\noindent\textbf{Turn 8: Termination}\\
\noindent\textit{User: "No, I think I have a clear picture now. Thanks for your help!"}\\
\noindent\texttt{System:} "You’re welcome! Best of luck with your bakery in San Ysidro. Feel free to reach out if you have more questions!"
\end{mdframed}
\caption{The simplified shows examples an ideation conversation with goals, subgoals and shifting intent. For lack of space, we do not elaborate on the subgoals.}
\label{fig:dialogue}
\end{figure*}
We illustrate an ultra-simplified 5-turn conversation in Figure \ref{fig:dialogue} between a MISCON user who is consulting the agent about starting a bakery business in San Ysidro, California. In Turn 1, the user species the intent of starting a bakery and a proposed location of the business. In response, the system consults its knowledge base to identify potential intentions that could be associated with businesses. In Turn 2, the user selects one of the suggested intents, and the system uses its demographic knowledge to identify challenges that the user should be aware of. In Turn 3, the user seeks a product line recommendation and the system makes minor suggestions and verifies that the user is aligned with the suggestion. In Turn 4, the user agrees and shifts focus to budgets, and in Turn 5, the conversation shifts again from budgets to permit requirements. Turn 6 highlights this shifting intent as the system consults its regulatory knowledge to identify that the fees depend on the layout of bakery, and the user details the layout of the bakery. While the system offers help with the submission of a permit application, the user shifts focus to finances and informs the system about his approved loan amount, and validates that it is adequate. Although the system offers to continue the conversation because it still has information to offer based on the conversation history, the user terminates the conversation in Turn 8, stating that his current goals are met. As in the previous turns, the system adjusts to the shift and closes the session.

\subsection{A State-space Model}
Thus, MISCON is designed as a state-space model to help users achieve specific objectives. However, user variability, external dependencies (e.g., legal or financial prerequisites), and the dynamic nature of conversational goals introduce challenges in designing such systems. We propose a framework that incorporates the following.
\begin{enumerate}[label=\arabic*.]
    \item Adaptive intent management to align with user states and milestones.
    \item External and internal milestones for structured progression.
    \item Adaptive knowledge-driven queries for information retrieval.
    \item Dynamic prioritization of milestones and entities.
\end{enumerate}

\begin{figure*}[t]
    \centering
    \begin{mdframed}
    \begin{verbatim}
Intent hierarchy (I):
1. permits and licenses
 1.1. Obtain a food facility health permit
 1.2. Obtain a business license
 1.3. Obtain any additional permits required by the city or county
2. Find a suitable location
 2.1. Determine the ideal neighborhood or area for the bakery
 2.2. Identify potential commercial properties for lease or purchase
 2.3. Evaluate the zoning requirements and restrictions for each property
3. Secure financing
 3.1. Determine the startup costs and operating expenses
 3.2. Explore financing options (e.g., small business loans, grants, investors)
 3.3. Prepare financial projections and loan applications
4. Develop a business plan
 4.1. Conduct market research and competitor analysis
 4.2. Define the bakery's concept, target audience, and unique value proposition
 4.3. Create a marketing and sales strategy
 4.4. Outline the operational plan and management structure
 4.5. Prepare financial projections and funding requirements
\end{verbatim}
\end{mdframed}
    \caption{A sample intent hierarchy that is pre-learned by the system.}
    \label{fig:intent-hierarchy}
\end{figure*}

In the following, we present the elements of the MISCON state-space conversational model against the backdrop of our simple example (Fig. \ref{fig:dialogue}).

\noindent\textbf{Conversation History ($H_C = \{u_1, v_1, u_2, v_2 \ldots \}$)} is an alternating sequence of user and system utterances. Each utterance $u_i$ (resp. $v_i$) has an utterance id $id(u_i)$, a timestamp $ts(u_i)$, the speaker id $uid(u_i)$ and the utterance text $t(u_i)$. 

\noindent\textbf{Goals ($G$)}
The agent's objectives are represented as a set of goals:$G = \{g_1, g_2, \dots, g_n\}$ where each $g_i$ is associated with a set of subgoals ($M$) and external dependencies ($M_{external}$). For a user exploring the feasibility of starting a bakery in San Ysidro, CA:
\begin{itemize}
    \item $g_1$: Explore market potential.
    \item $g_2$: Define product offerings.
    \item $g_3$: Assess financial feasibility.
    \item $g_4$: Identify regulatory requirements.
\end{itemize}

\noindent\textbf{Subgoals (Internal Milestones) ($M = \{m_1, m_2, \dots, m_k\}$)} guide the conversation toward achieving goals:
Each milestone $m_i$ is defined as:
\begin{equation}
m_i = (Description, P(E), Progress, Priority),
\end{equation}
where $P(E)$ represents the set of prerequisites (entities or conditions) required to achieve the milestone. The subgoals are specific target information that must be obtained from the user or the knowledge graph. For example, some subgoals can be:
\begin{itemize}
    \item $m_1$: Define business type and location ($P(E)$ = \{"Bakery", "San Ysidro"\}).
    \item $m_2$: Identify target market ($P(E)$ = \{"Demographics", "Spending Patterns"\}).
    \item $m_3$: Determine product offerings ($P(E)$ = \{"Product List"\}).
    \item $m_4$: Assess financial feasibility ($P(E)$ = \{"Rental Costs", "Pricing"\}).
    \item $m_5$: Complete regulatory steps ($P(E)$ = \{"Permits", "Licenses"\}).
\end{itemize}

\noindent\textbf{External Milestones ($M_{\text{ext}}$)} are external factors of the conversation that influence the user’s ability to progress. These can be divided into the following:
\begin{enumerate}[label=\alph*.]
    \item \textbf{User State Milestones}, such as the user has already obtained a necessary certificate.
    \item \textbf{Business Milestones}, such as a bakery may need to obtain Processed Food Registration (PFR) may be required from the California Department of Public Health.
\end{enumerate}
As mentioned earlier, user state milestones include factors like $m_{u_1}$: Work authorization and $m_{u_2}$: Financial readiness while business milestones include factors like $m_{b_1}$: Business registration and $m_{b_2}$: Permits and licenses. Note that while business milestones maybe incorporated into conversation subgoals, they are independent of the conversation because the user may achieve these milestones by himself.

\noindent\textbf{States ($S$)}
Conversation states $S = \{s_1, s_2, \dots, s_m\}$ capture the current context and progress that dynamically change at each time step, where each state $s_i$ is defined as: $s_i = (c_i, p_i, e_i)$ 
where $c_i$ is the context, $p_i$ represents progress and $e_i$ captures external milestone states. Thus, for our case, the states may represent
\begin{itemize}
    \item $s_1$: Initial inquiry ("I want to explore if starting a bakery in San Ysidro is feasible.")
    \item $s_2$: Market analysis ("What are the demographics and spending patterns in the area?")
    \item $s_3$: Definition of product ("What products are popular in this market?")
    \item $s_4$: Feasibility check ("What are the costs and competition?")
    \item $s_5$: Regulatory assessment ("What permits do I need?")
\end{itemize}

\noindent\textbf{State Transitions ($T = \{t_1, t_2, \dots, t_p\}$)} where each transition $t_i = (s_a, s_b, \phi, \alpha)$ is characterized by:
\begin{itemize}
    \item $s_a, s_b$: The start and end states.
    \item $\phi$: The trigger condition (e.g., user input, milestone completion, history context).
    \item $\alpha$: The system action required for the transition.
\end{itemize}

Examples of state transitions, updated with histories:
\begin{itemize}
    \item $t_1: s_1 \rightarrow s_2$ when the user asks about the market \textbf{and} $H_I$ shows no prior queries related to market demographics.
    \item $t_2: s_2 \rightarrow s_3$ when demographics are identified \textbf{and} $H_E$ confirms the necessary entities (e.g., target audience, spending patterns) are extracted.
    \item $t_3: s_3 \rightarrow s_4$ when products are defined \textbf{and} $H_C$ indicates that product-related discussions are complete.
    \item $t_4: s_4 \rightarrow s_5$ when costs and competition are assessed \textbf{and} $H_{M_{\text{ext}}}$ shows relevant external milestones (e.g., financial readiness) are satisfied.
\end{itemize}

\noindent\textbf{Entities and Events ($E = \{e_1, e_2, \dots, e_l\}$)}
Entities and events extracted from user utterances provide the conversational context. Each entity $e_i$ is defined by:
\begin{itemize}
    \item \textbf{Type}: The category of the entity (e.g., location, funding).
    \item \textbf{Value}: The specific value of the entity (e.g., ``San Ysidro,'', ``\$80,000'').
    \item \textbf{Priority}: A dynamically adjusted importance score.
\end{itemize}
In our example, examples of entities relevant to our example would be:
(\textit{Location = "San Ysidro", BusinessType = "Bakery", Demographics = "Young Families", ProductType = "Pan Dulce",  Pricing = "\$28-\$32" per dozen})

\noindent\textbf{Intents ($I = \{i_1, i_2, \dots, i_p\}$)}
represent the underlying purpose of the user, as estimated by the MISCON system. Each intention $i_i$ is timestamped and dynamically updated based on the user state derived from the conversation and the progression of the milestones. Intents are sometimes expressed as hierarchies, as shown in Figure \ref{fig:intent-hierarchy}

\noindent\textbf{Knowledge Base ($K$)}
The knowledge base contains domain-specific facts and rules, supporting efficient querying and adaptive information retrieval. Knowledge content covers attributes such as demographic data, current businesses, consumption patterns, and regulatory requirements in and around the San Ysidro area. We do not cover the details of the knowledge graph in this paper, but we show that the knowledge graph can be queried via query operations $q_1, q_2, \ldots$.

The behavior of the conversation system is governed by a set of operations over the set $G, S, M, M_{ext}, T, E, I, K$. The following shows the signature of these operations. The current implementation of these functions is briefly discussed in the next section. \\

\noindent\textbf{Intent Recognition ($I \times U \to i$)}
Maps a user utterance $U$ and the user conversation history $H_C$ to an intent $i \in I$ (e.g., recognizing intent in a user statement like ``I like where this is going'' needs the conversation history):
\begin{equation}
RecognizeIntent(U, I, H_C) = \arg\max_{i \in I} Score(U, i, H_C).
\end{equation}

\noindent\textbf{Entity Extraction ($U \to P(E)$)}
Extracts entities and events from user utterances:
\begin{equation}
P(E) = ExtractEntities(U,H).
\end{equation}

\noindent\textbf{ExtractMilestones ($U, S, H_C, H_{M_{\text{ext}}} \to M_{\text{new}}$)}\\
Identifies new external milestones from user input during a conversation:
\begin{equation}
M_{\text{new}} = ExtractMilestones(U, S, H_C, H_{M_{\text{ext}}}).
\end{equation}
The parameters to this function are:
\begin{itemize}
    \item \textbf{\(U\)}: The current user input, providing the primary source of information for milestone extraction.
    \item \textbf{\(S\)}: The current conversation state, including progress, known entities, and goals, for contextual analysis, e.g., when the user states``I have secured a loan for \$80,000."
    \item \textbf{\(H_C\)}: The conversation history, supplying prior user utterances for context and disambiguation.
    \item \textbf{\(H_{M_{\text{ext}}}\)}: The history of previously identified external milestones, ensuring no redundancies in the output.
\end{itemize}

\noindent\textbf{Output:}
\begin{itemize}
    \item \textbf{\(M_{\text{new}}\)}: A set of newly extracted external milestones that are not yet part of \(H_{M_{\text{ext}}}\).
\end{itemize}

\noindent\textbf{ConstructQueryGraph ($I \times P(E) \to Q$)}
Constructs a query graph based on the recognized intent and extracted entities:
\begin{equation}
Q = ConstructGraph(I, P(E)).
\end{equation}

\noindent\textbf{RetrieveFacts ($Q \times K \to F$)}
Retrieves facts from the knowledge base based on the query graph:
\begin{equation}
F = QueryKnowledgeBase(Q, K).
\end{equation}

\noindent\textbf{GenerateResponse ($F \times S \to R$)}
Generates a system response:
\begin{equation}
R = GenerateResponse(F, S).
\end{equation}

\noindent\textbf{findMissingInformation ($S \times P(E) \to P(E)$)}
Identifies missing entities or conditions:
\begin{equation}
P(E)_{missing} = findMissingInformation(S, P(E)).
\end{equation}

\noindent\textbf{getClarification ($P(E) \to C$)}
Generates clarification questions:
\begin{equation}
C = GenerateClarification(P(E)_{missing}).
\end{equation}

\noindent MISCON also has a set of additional functions to adapt to the variations and drifts that might appear in the conversation. Every updated element below has an associated update time.\\

\noindent\textbf{Milestone Prioritization}
The priority of a milestone $m_i$ is calculated as:
\begin{multline}
Priority(m_i) = w_{progress} \cdot P_{progress} + w_{relevance} \cdot \\ P_{relevance} + w_{external} \cdot P_{external}
\end{multline}\\
where $w_{progress}, w_{relevance}, w_{external}$ are weights that depend on the context of the conversation.

\noindent\textbf{UpdateEntityPriorities ($P(E) \times S \to P(E)$)}
Adjusts entity priorities dynamically:
\begin{equation}
P(E)' = UpdatePriorities(P(E), S).
\end{equation}

\noindent\textbf{Dynamic Intent Management}
Intents are dynamically updated based on the user state and milestone progression:
\begin{equation}
I' = UpdateIntents(I, S).
\end{equation}

\noindent\textbf{Adaptive Queries}
Queries adapt to the user’s state and context. Currently, this is performed via a set of rules. For example, a query can be 
\begin{equation}
Q = 
\begin{cases}
ExpandQuery(Q) &\texttt{if}~ est(.) < \tau, \\
RefineQuery(Q) &\texttt{otherwise}.
\end{cases}
\end{equation}
where estimation predicate $est(.)$ combines the user's prior contextual knowledge based on the registration-time information (e.g., educational level) and conversation history, together with information provided in the conversation history. 

\begin{algorithm*}
\caption{The conversation algorithm of MISCON with incremental external milestones}
\label{algo:basic}
\begin{algorithmic}[1]
\State Initialize $G, S, M, M_{\text{ext}}, T, K$.
\State Initialize histories $H_C, H_I, H_{M_{\text{ext}}}, H_E \gets \emptyset$.
\While{Conversation is active}
    \State Receive user input $U$.
    \State Append $U$ to $H_C$.

    \State \textbf{Update External Milestones:}
    \State Extract new milestones from user input: $M_{\text{new}} \gets \text{ExtractMilestones}(U)$.
    \State Update external milestones: $M_{\text{ext}} \gets M_{\text{ext}} \cup M_{\text{new}}$.

    \State \textbf{Perform External Milestone Checks:}
    \ForAll{$m_{\text{ext}} \in M_{\text{ext}}$}
        \If{$m_{\text{ext}}$ unresolved \textbf{and} $m_{\text{ext}} \notin H_{M_{\text{ext}}}$}
            \State Generate clarification prompts $C_{\text{ext}} \gets \text{getClarification}(m_{\text{ext}})$.
            \State Send $C_{\text{ext}}$ to User Interface.
            \State Append $m_{\text{ext}}$ to $H_{M_{\text{ext}}}$.
        \EndIf
    \EndFor

    \State Perform Intent Recognition: $i \gets \text{IntentRecognition}(U, I)$.
    \State Append $i$ to $H_I$.

    \State Perform Entity Extraction: $P(E) \gets \text{EntityExtraction}(U)$.
    \State Append $P(E)$ to $H_E$.

    \State Update state $S$ using $i, P(E), H_C, H_I, H_E$.

    \State \textbf{Check for Missing Information:}
    \State $P(E)_{\text{missing}} \gets \text{findMissingInformation}(S, P(E))$.
    \If{$P(E)_{\text{missing}} \neq \emptyset$}
        \State Generate clarification $C \gets \text{getClarification}(P(E)_{\text{missing}})$.
        \State Send $C$ to User Interface.
        \State Append $C$ to $H_C$.
        \State \textbf{Continue}.
    \EndIf

    \State Construct query graph $Q \gets \text{ConstructQueryGraph}(i, P(E))$.
    \State Retrieve facts $F \gets \text{RetrieveFacts}(Q, K)$.
    \State Generate response $R \gets \text{GenerateResponse}(F, S)$.
    \State Append $R$ to $H_C$.

    \State Update priorities $P(E)' \gets \text{UpdateEntityPriorities}(P(E), S)$.
    \State Update intents $I' \gets \text{UpdateIntents}(I, S)$.

    \State \textbf{Adapt Query Using User Metadata:}
    \State Decision $D \gets \text{est}(S, \text{UserMetadata})$.
    \State Adapt query $Q \gets \begin{cases}
        \text{ExpandQuery}(Q) & \text{if } D = \text{``expand''}, \\
        \text{RefineQuery}(Q) & \text{if } D = \text{``refine''}, \\
        \text{NoChange}(Q) & \text{if } D = \text{``no change''}.
    \end{cases}$

    \State \textbf{Check Goal or Termination:}
    \If{$G$ achieved \textbf{or} user ends conversation}
        \State Terminate.
    \EndIf
\EndWhile
\end{algorithmic}
\end{algorithm*}

\section{Implementing the Model}

Algorithm \ref{algo:basic} shows the basic structure of conversation system that implements the model. A comprehensive description of the implementation of all functions in the algorithm is beyond the scope of this paper. More importantly, at the time of writing this work-in-progress report, we are still experimenting with the implementation of the functions outlined above, to achieve a balance between \textit{factual accuracy}, \textit{user-level interpretability}, \textit{responsiveness} in terms of interactivity and \textit{ cost effectiveness} in terms of using our own algorithms, fine-tuning an LLM like Llama 3.2, directly using GPT 4o \cite{foosherian2023enhancing}. In the following, we identify a few of the decisions and design trade-offs that we consider.\\

\noindent \textbf{Factual Accuracy.} For our experimental domain of San Diego and Imperial Counties, we have constructed a heterogeneous collection of documents (e.g., municipal, county, state and food-related federal laws/ordinances and permits), geospatial regional (most recent and annually updated survey data at the level of block groups) and point (business locations) information, ontologies (e.g., FOODON, the food ontology), non-ontological graph data (food-product, ingredient, nutrition, recipe graphs), web-extracted semi-structured data from organizations (e.g., nonprofits) belonging to the food system or public service organizations. These diverse data are served through the AWESOME polystore \cite{dasgupta2016analytics,zheng2023optimized}, and a virutal knowledge graph constructed on top of the AWESOME system \cite{zheng2023p2kg,desai2024building}. The $ConstructGraph$ function performs a schema retrieval operation to identify the portion of the knowledge graph that can satisfy a query and auto-generates a query against the system; the $QueryKnowledgeBase$ operation issues the query against the virtual knowledge graph and $RetrieveFacts$ composes a semi-structured result (e.g., containing demographic data as well as regulatory information) to return to the conversation. The $generateResponse$ function uses an LLM to transform this result into several ``bite-sized'' pieces of textual responses that are returned to the user for better interpretability. For the final result delivery phase of the $GenerateResponse$ operation, we use an LLM to paraphrase the results according to the system's estimation of the user's comprehension capabilities.\\

\noindent \textbf{User-level Interpretability.} We consider a system response to be ``interpretable'' when it addresses both the \textit{conversational intent} and the \textit{domain question intent} at the estimated level of user comprehension. A conversational intent roughly corresponds to the intent class of the utterance. For example, factual questions, recommendation request, opinion, comparison request, and disagreement are a few of the 22 categories of conversational intents we consider. Thus, a factual question about a business location and the recommendation request about a business location are treated differently. A domain intent, on the other hand, is about the expected content of the system's response. For example, for the larger task of business ideation, we have identified about 300 different food business types and have collected a total of about 44000 user intents. Many of them are common to multiple businesses as shown in The intention tree in Figure \ref{fig:intent-hierarchy}, while others, not shown, are very specific to the type of business. Although the $RecognizeIntent$ function uses the conversation history $H_C$ to ensure more accurate intent estimation, the optimal context to be used is still under research.  

A different dimension of interpretability arises from the system's estimate of the user's readiness level for an answer. For example, often complex legal clauses need to be simplified, statistical definitions (e.g., the average vs. the median income) must be described, and recommendations must be broken down into pros and cons. We can infer the conversational intent with an F1 score of over 0.9. We are currently trying to evaluate how fine-tuned pretrained LLMs perform in terms of result quality and interpretability.  \\

\noindent \textbf{Responsiveness.} Responsiveness is at the heart of any conversation system because the human user expects the AI agent to respond with interactive speed. Querying the knowledge graph becomes a cost-dominant operation when the system needs to generate a large cascade of internal operations (i.e., a series of $QueryKnowledgeBase$ calls) to produce a good-quality response (currently evaluated by our human testers). A hint of the problem is shown in Turn 2 of the conversation in Figure \ref{fig:dialogue}. The expanded version of the market suitability question first breaks down the suitability based on multiple axes including income distribution and spending patterns, ethnic composition and food preference, competition, recent changes in the neighborhoods, sends multiple queries to the knowledge graph to find all partial answers, combines them, sends the composite response to multiple LLMs and assimilates their responses. The higher the number of such accesses and the complexity of the knowledge graph for each such axis, the longer it takes to produce a response. We are exploring how partial responses (e.g., competing businesses) generated by the knowledge graph can be returned to the user as they are computed.\\

\noindent \textbf{Cost Effectiveness.} We use the term \textit{cost} to represent the monetary expense in making LLM calls. We distinguish between two different cost components: $C_1$, the upfront LLM costs that are incurred for fine-tuning and developing high-quality prompt libraries, and $C_2$,  costs incurred for LLM calls made during a conversation session per user. While $C_1$ is higher and is proportional to the size of the training data and the complexity of tasks, $C_2$ is a recurring cost and will dominate over $C_1$ over time \cite{lee2023platypus}. 

In the current implementation, different LLMs are used within different operations via task-specific prompt templates where MISCON's employs a \textit{result validation process} and its own \textit{graph-structured cache} to maintain and accumulate results from partial results obtained through incremental prompting. For example, we have determined that the extraction of entities, events, and their semantic classes is better with GPT 40-mini compared to any traditional NER solution or a non-commercial LLM like Llama 3.2 70B primarily because of the ``openness'' of the conversation domain we referred to earlier. At the same time, switching to GPT 4o, does not provide substantively better results. We take a more nuanced approach for a task like intent recognition. Although recognition of conversational intent is achieved easily with fine-tuned BERT, recognition of domain intent is harder. If the user's intent closely matches the domain intent tree via embeddings computed from our training set, the results are usually accurate. However, if predicted intent does not fall within a small distance of the node vectors of the intent tree, the results are often vastly inaccurate. In this case, using GPT 4o, although more expensive, produces much better results compared to a fine-tuned Llama or even GPT 4o mini. An in-depth study of the comparative abilities of different LLMS towards these knowledge tasks is part of our future work.\\

\noindent \textbf{Pragmatics.} We enforce domain constraints by validating that an user utterance directly or contextually refers to the food business. To assess if the conversation is drifting out of domain, we currently use (i) a topic coherence metric measured by the proportion of new entities are introduced in the current utterance compared to the entities referred to before, and (ii) whether the conversation does not progress because the utterances are repeating the same set of states for several turns. In our example, the first case would occur if the user suddenly starts inquiring about a catering business, and the second case would occur if the user if a user says ``Tell me again how permits affect my budget?" In cases like this, the system detects and documents the conversation and refers the user back to a human advisor.

\section{Conclusion}

With MISCON in NOURISH, we are trying to achieve two different end results -- creating an actual user-facing conversation system that provides them with real, useful consulting for food businesses in nutrition-insecure places, and  creating a state-of-the-art functional research platform that can identify, characterize, and improve upon knowledge-based multi-session dialogue systems that persistently maintain user states. In the process, we are discovering many methodological hurdles, unexplored opportunities, and trade-offs that are inherent to semi-open-domain conversation systems that need a large and heterogeneous knowledge base and are not commonly seen in the research literature. 

\bibliography{aaai25}

\end{document}